\documentclass[sigconf,anonymous=false]{acmart}
\renewcommand\footnotetextcopyrightpermission[1]{} % removes footnote with conference information in first column
\usepackage{booktabs} % For formal tables
\usepackage{enumitem}
\usepackage{multirow,array}
\usepackage{tikz}
\usepackage{amsmath}
\usepackage{float}
\usepackage{breqn}
\usetikzlibrary{trees}
\usepackage{tabularx}
\usepackage{microtype}

\setcopyright{rightsretained}
\acmDOI{10.475/123_4}

\acmISBN{123-4567-24-567/08/06}

 \acmConference[ACAN'18]{11th International Workshop on Automated Negotiations}{July 2018}{Stockholm, Sweden}
\acmYear{2018}
\copyrightyear{2018}

\acmBooktitle{Proceedings of the 11th International Workshop on Automated Negotiations (held in conjunction with IJCAI 2018)}
\begin{document}
\title[Agents with different behaviors for Contract Negotiation using Reinforcement Learning]{ Prosocial or Selfish? Agents with different behaviors for Contract Negotiation using Reinforcement Learning}
%\titlenote{Produces the permission block, and
%  copyright information}
%\subtitle{Extended Abstract}
%\subtitlenote{The full version of the author's guide is available as
%  \texttt{acmart.pdf} document}

\author{Vishal Sunder}
\affiliation{%
  \institution{TCS Research}
  \city{New Delhi}
  \country{India}
}
\email{s.vishal3@tcs.com}

\author{Lovekesh Vig}
\affiliation{%
  \institution{TCS Research}
  \city{New Delhi}
  \country{India}
}
\email{lovekesh.vig@tcs.com}
\author{Arnab Chatterjee}
\affiliation{%
  \institution{TCS Research}
  \city{New Delhi}
  \country{India}
}
\email{arnab.chatterjee4@tcs.com}
\author{Gautam Shroff}
\affiliation{%
  \institution{TCS Research}
  \city{New Delhi}
  \country{India}
}
\email{gautam.shroff@tcs.com}

% The default list of authors is too long for headers.
\renewcommand{\shortauthors}{Vishal et al.}

\begin{abstract}
We present an effective technique for training deep learning agents capable of negotiating on a set of clauses in a contract agreement using a simple communication protocol. We use Multi Agent Reinforcement Learning to train both agents simultaneously as they negotiate with each other in the training environment. We also model selfish and prosocial behavior to varying degrees in these agents. Empirical evidence is provided showing consistency in agent behaviors. We further train a meta agent with a mixture of behaviors by learning an ensemble of different models using reinforcement learning. Finally, to ascertain the deployability of the negotiating agents, we conducted experiments pitting the trained agents against human players. Results demonstrate that the agents are able to hold their own against human players, often emerging as winners in the negotiation. Our experiments demonstrate that the meta agent is able to reasonably emulate human behavior.
\end{abstract}

%
% The code below should be generated by the tool at
% http://dl.acm.org/ccs.cfm
% Please copy and paste the code instead of the example below.
%
%\begin{CCSXML}
%<ccs2012>
% <concept>
%  <concept_id>10010520.10010553.10010562</concept_id>
%  <concept_desc>Computer systems organization~Embedded systems</concept_desc>
%  <concept_significance>500</concept_significance>
% </concept>
% <concept>
%  <concept_id>10010520.10010575.10010755</concept_id>
%  <concept_desc>Computer systems organization~Redundancy</concept_desc>
%  <concept_significance>300</concept_significance>
% </concept>
% <concept>
%  <concept_id>10010520.10010553.10010554</concept_id>
%  <concept_desc>Computer systems organization~Robotics</concept_desc>
%  <concept_significance>100</concept_significance>
% </concept>
% <concept>
%  <concept_id>10003033.10003083.10003095</concept_id>
%  <concept_desc>Networks~Network reliability</concept_desc>
%  <concept_significance>100</concept_significance>
% </concept>
%</ccs2012>
%\end{CCSXML}

%\ccsdesc[500]{Computer systems organization~Embedded systems}
%\ccsdesc[300]{Computer systems organization~Redundancy}
%\ccsdesc{Computer systems organization~Robotics}
%\ccsdesc[100]{Networks~Network reliability}

%\keywords{ACM proceedings, \LaTeX, text tagging}

\maketitle

\section{Introduction}
\label{sec:intro}
Reaching a consensus on the contents of a contract agreement is often an expensive and time consuming task for the negotiating parties involved. Automating this process is important not just from an industrial point of view but also for all related  problems where negotiation is required. The primary challenge in this endeavor is to suitably model the behavior of an \textit{AI agent}. Several papers in the literature \cite{cao2015automated,hou2004modelling} have addressed the problem of modeling agents for contract negotiation. However, most of the proposed approaches use a combination of mathematical models and rules. With the upsurge of \textit{deep reinforcement learning}, in the past few years, it becomes interesting to analyze the behavior learnt by \textit{deep} agents trained to negotiate via reinforcement learning. Recently, we have seen valuable contributions in the domain of \textit{AI agents'} negotiation using deep reinforcement learning \cite{cao2018iclr,lewis2017deal} but much work needs to be done for their direct applicability to various negotiation oriented domains such as contract negotiation.

Although there is training data available for a negotiation domain in which agents learn to divide items between themselves \cite{lewis2017deal} (this is the problem of resource reallocation which has been the formalism in most negotiation studies), no such data is available for contract negotiation where agents need to reach a consensus about which items to share between them (clauses). This makes it difficult to train an agent which can converse in natural language. Although crowdsourcing can be utilized for generating data but it is resource intensive. Nevertheless, we present an approach where we train agents to converse with each other using a simple protocol which is an interpretable sequence of bits. This protocol is scalable and robust. The training is done using reinforcement learning. We first model an agent as a neural network and then train two such agents concurrently where they play several rounds of negotiation games against each other and  learn to coordinate with each other based on the outcome (reward). The behavior of the agents is modeled by using the effective technique of varying the reward signal. In this way, we get agents with different behaviors. We show that agents trained in this manner indeed learn to coordinate their moves and produce context relevant outputs. We observe that agents have consistency in their behaviors and this is shown by making all agents negotiate against each other and analyzing the results.

For emulating human behavior, we train an agent (\textit{meta agent}) with a dynamic behavior. We use a model which is an ensemble of the 4 agents mentioned above, with a "selector" agent on top which selects the appropriate behavior based on the negotiation state. This agent is also trained using reinforcement learning. We have found evidence that the meta agent learns a simple decision tree while selecting it's behavior.

Finally, for assessing the usability of the negotiation agents in real world scenarios, we conduct experiments where we evaluate the agents against human players. We show that agents show consistency in behaviors even against human players. Thus, we show that our agents are deployable in real industrial scenarios for negotiating on a contract.

Hence, the main contributions of this paper are:
\begin{enumerate}
\item A deep learning model and a reinforcement learning procedure for training an AI agent to negotiate in the domain of contract negotiation.
\item Modeling selfish/prosocial behavior by varying the reward signal for the agent and its opponent in the reinforcement learning framework and empirical evidence for the same.
\item An AI agent with a dynamic behavior (varying within a negotiation instance) by learning an ensemble of different agent behaviors using reinforcement learning.
\item Evidence for the usability and success of the negotiation agent against human players through real life experimental results. 
\end{enumerate}
The rest of the paper is organized as follows: Section~\ref{sec:prob_form} formulates the problem statement and provides a general framework for our negotiation environment. It also describes the model for the negotiation agent and  the evaluation metrics. Section~\ref{sec:learn_pr} describes the model architecture and learning procedure. Section~\ref{sec:coord} analyzes the results when the agents which are trained against each other negotiate. Section~\ref{sec:interplay} provides an analysis of the behavior of the agents and demonstrates consistency in the learnt behaviors. Section~\ref{sec:model_mix} provides details about  the model and performance of the \textit{meta agent} which has a mixed/dynamic behavior. Section~\ref{sec:human_eval} gives the results of the human evaluation. Finally, we conclude and give directions for future work in Section~\ref{sec:conc_fw}.

\section{Related Work}
\label{sec:relw}
Contract negotiation can be seen formally as a resource reallocation problem~\cite{dunne2005extremal,dunne2005complexity}. Several papers have addressed the problem of building AI agents that can negotiate on a contract. But most have focused on domain specific applications of the same~\cite{yu2012financial,camarinha2007contract}. However, our approach to this is one where we require agents to reason about a shared set of decision variables (clauses)~\cite{faratin2001automated}. The problem formulation is similar to the furniture layout task~\cite{english2005learning,yang2004using}, where two conversants try to agree on five pieces of furniture to place in a room.

In general, much work on building negotiation agents has focused on mathematically modeling these agents or building fine grained rules~\cite{camarinha2007contract,cao2015automated,zuzek2008formal,sierra1999service}. However, machine learning has seldom been used for tackling this problem. One of initial works on the use of Reinforcement Learning in negotiation focused on formally defining a reinforcement learning state for the task~\cite{heeman2009representing}. 

\textit{Policy Hill Climbing (PHC)} has been shown to converge better than traditional algorithms such as Q-Learning~\cite{georgila2014single} for multi agent training, but convergence becomes an issue with an increase in size of the \textit{action space}. This is where modern \textit{Deep Learning} methods become useful.

With recent advancements in deep learning, there has been a lot of interest in building agents that learn to communicate and coordinate for various multi agent tasks. Deep reinforcement learning methods can be used to train agents to cooperate in complex environments~\cite{lerer2017maintaining}. Furthermore, it is possible to build deep learning agents that can be trained to develop their own protocols to solve collaborative tasks~\cite{foerster2016learning,cao2018iclr}.

More recently, we see valuable contributions in building \textit{deep} agents for negotiation domain. With the availability of data, deep learning agents can be trained to imitate humans (converse in natural language) and subsequently even outperform them by using reinforcement learning techniques~\cite{lewis2017deal}. Although robust and end to end, such models require training data, collecting which can be resource intensive for different domains.

Modeling negotiation behavior by varying the reward signal has been used previously in Ref.~\cite{cao2018iclr}, but their main aim is to study the communication protocols that can emerge during negotiation. They limit themselves to two behaviors whereas we explore four behaviors in addition to a dynamic behavior agent.

With the exception of Ref.~\citep{lewis2017deal}, none of the previous works, to the best of our knowledge provide qualitative/quantitative results against human players.

\section{Problem Formulation}
\label{sec:prob_form}
Here, we formulate the problem of negotiating on a set of clauses in a contract agreement scenario wherein two parties have to reach a consensus as to which clauses need to be included in / excluded from the final draft. For this paper, we have considered $6$ clauses but our method is generalizable for any number of clauses (We have considered $6$ clauses because we found that this reduces the training time of our model while preserving enough complexity in the problem). 
\subsection{Negotiation Environment}
\label{sec:nego_env}
The value that an agent attaches to the clauses is represented by a utility function which is a vector of 6 integers between -12 and 12 (excluding 0) such that their sum is zero. There is an additional constraint that there is at least one positive and one negative value in this vector and that the sum of positives is $+12$ and that of the negatives is $-12$. More formally, this vector is represented as:
\begin{align}
U=\textrm{Shuffle}(P \bigoplus N)
\end{align}
\begin{align}
P=[p_1,p_2,...,p_k]
\end{align}
\begin{align}
N=[n_1,n_2,...,n_{6-k}]
\end{align}
where,
$0<k<6$, $\bigoplus$ is the concatenation operator and Shuffle(.) is a "random shuffling" function. 
Also, $p_i \in \{1,...,12\}$ and $n_i \in \{-12,...,-1\}$, along with the following constraints:
\begin{align}
\sum_{i}p_i = 12
\end{align}
\begin{align}
\sum_{i}n_i = -12
\end{align}

Each element in the list represents the "importance" that the agent attaches to the corresponding clause. We have this distribution so that in every case there is a mixture of the most beneficial clauses (values summing to $12$) and the most harmful clauses (values summing to $-12$). For example, if an agent has a utility $[9, -5, 2, -1, -6, 1]$, clause one is non negotiable as it has the highest value. On the other hand, clauses three, four and six might be negotiable as they do not have a high absolute value (Of course, the decision as to whether a clause is negotiable or not has to be made the the agent). Notice that the maximum points an agent can score is $12$ and the minimum is $-12$. We exclude the "zero-valued" (neutral) clause to avoid skewness of values in the utility function.

Each of the agents receive this utility which is sampled uniformly. The agents then communicate with each other by giving offers, which is a sequence of 6 bits, $S_t \in \{0,1\}^6$. Here, subscript $t$ refers to the time-step at which the offer was produced. Each bit in this sequence is the agent's decision on the corresponding clause (0 meaning exclude and 1 meaning include). The communication follows a sequential structure and the agent that goes first is decided by an unbiased coin flip. This communication goes on between agents until one of the following happens:
\begin{enumerate}
\item An agreement is reached. This happens when an agent gives the same offer that it receives.
\item The time runs out. We keep a limit of 30 offers (15 for each agent) after which the negotiation process stops with a disagreement\footnote{We have run our experiments for different number of clauses and thresholds and have found that the agents require a certain minimum number of turns during training so that they learn to reach an optimal solution. For 6 clauses this turns out to be 30. A more thorough analysis of the relation between number of clauses and the threshold may form an interesting subject for future work.}.
\end{enumerate}

At the end of the negotiation, each of the negotiating parties get a reward based on the agreed sequence of bits. So, if agents A and B have utilities $U^A$ and $U^B$ respectively, and the agreed sequence is $S$, A gets $S.U^A$ and B gets $S.U^B$, where $.$ represents the dot product.
\begin{figure}
\includegraphics[width=8cm]{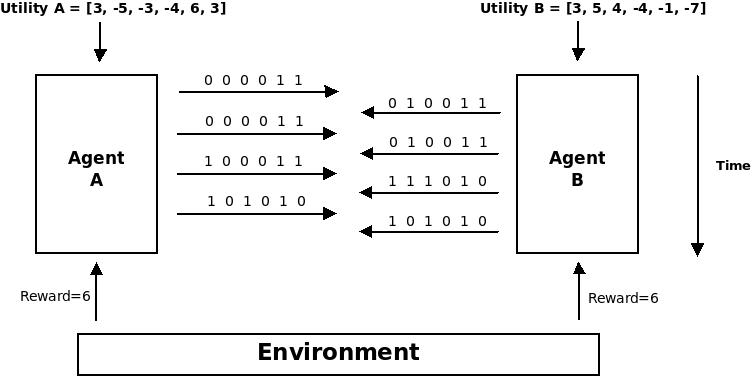}
\caption{A negotiation example.}
\label{fig:neg_eg}
\end{figure}
An example negotiation between two agents is illustrated in Figure~\ref{fig:neg_eg}.

\subsection{Negotiation Agent}
\label{sec:nego_ag}
Given a utility function and the opponent's offer, we divide the problem of producing an offer into two parts:
\begin{enumerate}
\item \textit{Deep Neural Network (DNN) Component}: The first part is to decide the number of bits to be flipped in the opponent's offer. This part is taken care of by the deep neural network which is trained through \textit{Reinforcement Learning}.
\item \textit{Rule Based Component}: Once the neural network decides the number of bits to be flipped, the part of deciding which exact bits to flip is deterministic i.e. flip the bits that result in maximum increase in score.

For example, if the utility is $[2, -6, -2, -4, 7, 3]$, the opponent's offer is $[1,1,1,0,0,1]$ and the number of bits to be flipped is $3$ (decided by the neural network), we would flip the second, third and fifth bit (rule based) as that would increase the score by $6 + 2 + 7 = 15$ points and we would get $12$ points (which is the maximum one can get).
\end{enumerate}

This approach has two distinct advantages:
\begin{enumerate}
\item The action space for reinforcement learning reduces significantly. This is particularly advantageous when the number of clauses are much higher than $6$. In general, for $n$ clauses, the action space goes from $2^n$ to $n$ and hence simplifies the training procedure.
\item The rule based component provides us leverage over the agent making it suitable for deployment against humans (see Section~\ref{sec:human_eval}). Without the rule based component, the deep network might learn policies which are suitable for negotiation with AI agents but not against humans~\cite{cao2018iclr,lewis2017deal}.
\end{enumerate}
We call this negotiation agent the \textit{RL Agent}.

\subsection{Evaluation Metrics}
\label{sec:eval}
The evaluation of a negotiation is done based on the following metrics:
\subsubsection*{Dialog Length} The dialog length is the average number of time steps for which the negotiation lasts.
\subsubsection*{Agreement Rate} Percentage of negotiations that end in an agreement represent the agreement rate.
\subsubsection*{Optimality rate} The optimality rate is the percentage of negotiations that end in an optimal deal. We say that a deal is optimal if it is both \textit{Pareto Optimal} \cite{raiffa1982art} and the both agents receive a positive score.
A solution is Pareto optimal if neither agent's score can be improved without lowering the other's score.
\subsubsection*{Average score} The average number of points earned by an agent. To see the maximum joint reward the agents can earn on an average on optimal deals, we greedily search through all possible deals ($2^6 = 64$) for all of the samples in the test set and select the one which results in the maximum joint reward and optimal deals. The average of maximum joint reward for the test set is $1.40$ ($0.70$ for each agent \footnote{We normalize the points earned by the maximum points possible i.e. 12.0}. This is used for comparison of average scores of trained agents in Table~\ref{tab:coord} and Table~\ref{tab:interplay}).

\section{Learning Procedure}
\label{sec:learn_pr}
We define below, the architecture and training procedure for the DNN component of the negotiation agent.
\subsection{Model Architecture and learning}
\label{sec:mdl_arch}

At a given time step $t$ in the negotiation process, an agent $A$ receives the following as the \textit{state} input:
\begin{enumerate}
\item It's Utility function $U^{A}$.
\item Offer given by opponent $B$, $S_{t}^{B}$.
\item It's previous offer, $S_{t-1}^{A}$.
\item Agent ID, $I \in \{0,1\}$.
\end{enumerate}
We convert this input into a dense representation $D_{t}^{A}$ as\begin{multline}
D_{t}^{A} = [\textrm{OfferMLP}([U^{A},S_{t}^{B}]), \textrm{OfferMLP}([U^{A}, S_{t-1}^{A}]), \\ \textrm{AgentLookup}(I), \textrm{TurnLookup}(t)].
\end{multline}

Here OfferMLP(.) is a 2-layer MLP with ReLU activation \cite{nair2010rectified}, AgentLookup(.) is an embedding which gives a dense representation for the agent identity and TurnLookup(.) is another embedding which encodes information in the timestep.

The representation $D_{t}^{A}$ is passed to a 2-layer GRU \cite{cho2014properties} as
\begin{align}
h_t^A = \textrm{GRU}(D_{t}^{A},h_{t-1}^A),
\end{align}
where $h_{t-1}^A$ is the hidden state generated by $A$ at its previous turn. The number of bits to be flipped are predicted (action taken) by sampling from policy $\pi_A$:
\begin{align}
\pi_A = \textrm{Softmax}(Wh_t^A).
\end{align} 
During test time though, we do not sample and select the action with the highest probability.

At the next time step, the agent $B$ also outputs a similar policy $\pi_B$. Each agent $i \in \{A,B\}$ tries to maximize the following objective individually:
\begin{align}
L_i = \mathop{\mathbb{E}}_{x_t \sim (\pi_A,\pi_B)}\bigg[\sum_{t}\gamma^{(T-t)}(r_i(x_{1...T}) - b_i)\bigg] + \lambda H[\pi_i].
\end{align}
Here,
\begin{enumerate}
\item $x_t$ is the action taken by an agent at time $t$,
\item $\gamma$ is the \textit{discount factor},
\item $T$ is the total time steps for which the negotiation lasts,
\item $r_i(x_{1...T})$ is the reward received by agent i at the end of the negotiation which is a function of the sequence of actions $x_t$ taken by the agent from $t=1$ to $t=T$,
\item $b_i$ is the baseline which is used to reduce variance, and 
\item $H[\pi_i]$ is the entropy regularization term to ensure exploration and $\lambda $ controls this degree of exploration \cite{mnih2016asynchronous}.
\end{enumerate}
The gradient of $L_i$ is estimated and the parameters updated as in REINFORCE \cite{williams1992simple}.

The parameters of agents $A$ and $B$ are shared. These parameters are updated after each episode. An episode refers to a negotiation game between agents $A$ and $B$. We run the training for $5$ epochs with $10^5$ episodes in each epoch.
The values of all hyperparameters is given in Appendix \ref{app:hyp}. 

\subsection{Behavior Modeling (Reward)}
\label{sec:behav_mod}

\begin{table}
  \caption{Behavior induced in agents A and B depending on whether there is an optimality signal in the rewards. The agents in the first two rows are subject to the same signal. Hence, they have similar tags in the brackets.}
  \label{tab:BM}
  \begin{tabular}{cccc}
    \toprule
    \multicolumn{2}{c}{Behavior} & \multicolumn{2}{c}{Signal} \\
    \midrule
     A & B & A & B \\
     \midrule
     Prosocial (Agent 1) & Prosocial (Agent 1) & Yes & Yes \\
     Selfish (Agent 2) & Selfish (Agent 2) & No & No \\
     Prosocial (Agent 3) & Selfish (Agent 4) & Yes & No \\
  \bottomrule
\end{tabular}
\end{table}

The manner in which rewards are given to the RL agent decides its behavior. In particular, we enforce \textit{selfish} and \textit{prosocial} behavior in the RL agent in the following way (although there may be a plethora of complex behaviors, we limit ourselves to the simplest ones):
\begin{enumerate}
\item To enforce prosocial behavior, the agent is given the reward (the number of points earned at the end of the negotiation) iff the deal is \textit{optimal}. If the deal is not optimal, the agent is given a reward of -0.5. This ensures that the agent not only cares about its own gain/loss while learning its policy but also takes into account the opponent's priorities as well. In other words, the reward here has a signal for the overall optimality.
\item If there is no optimality signal in the reward, i.e. the agent receives as a reward, whatever it earned in the negotiation, then a selfish behavior is induced. The agent then, learns to maximize its own score.  
\end{enumerate}
Both agents receive a reward of -0.5 if the negotiation ends in a disagreement.

As two agents learn concurrently, we get agents with 4 different behaviors depending on how the opponent is trained to behave (Table~\ref{tab:BM}).
\begin{enumerate}
\item \textit{Prosocial agent trained against a Prosocial agent (PP)}:
We get the behavior \textit{PP} when both agents are trained to have a prosocial behavior.

\item \textit{Selfish agent trained against a Selfish agent (SS)}: 
If both agents are trained to be selfish we get the agent \textit{SS}.

\item \textit{Selfish agent trained against a Prosocial agent and vice-versa (SP,PS)}: 
When one agent is trained to be selfish and its opponent is trained to be prosocial, we get two agents whom we call \textit{SP} and \textit{PS} respectively.
\end{enumerate}

\begin{figure}
\includegraphics[width=8.5cm]{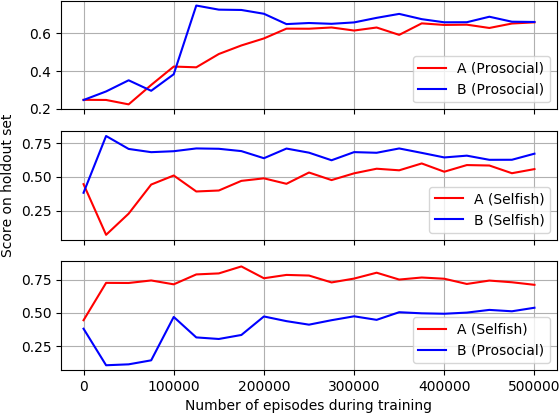}
\caption{Learning Curves.}
\label{fig:learn_cur}
\end{figure}

\begin{figure*}
\includegraphics[width=14cm]{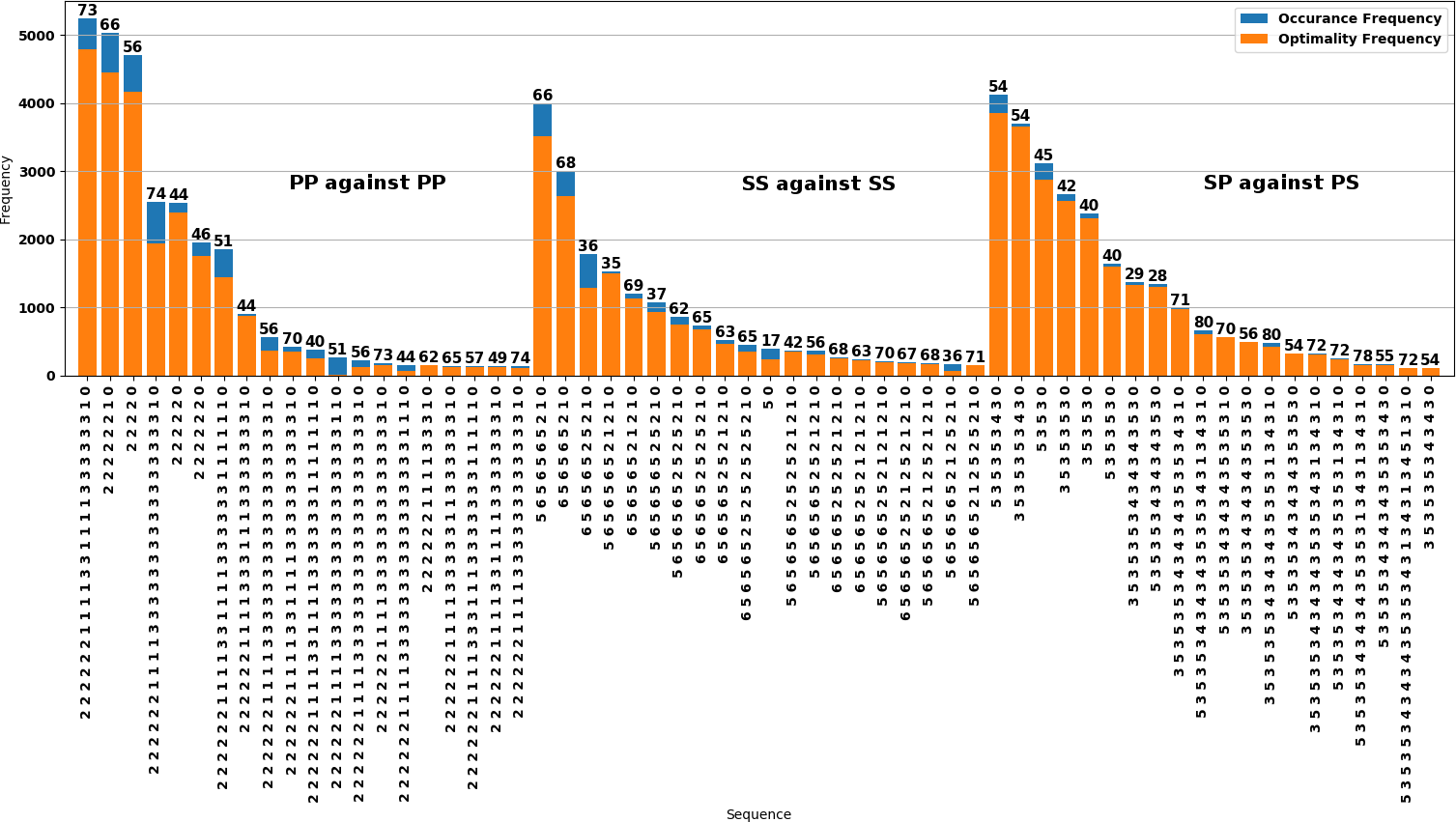}
\caption{Frequency distribution of the sequence of actions taken by the agents for all three cases. This only shows the top 20 occurrences. The orange part represents the number of optimal deals when the \textit{agents decide} to use the corresponding sequence. Each point on the x axis represents the sequence of number of bits flipped by the agents one after the other. The number on the top of each bar is the percentage optimality using the given sequence throughout.}
\label{fig:freq_dist_seq}
\end{figure*}

\section{Coordination between Trained Agents}
\label{sec:coord}

Here we analyze the results of the negotiation between agents who have been trained against each other. To assess whether these agents are learning to coordinate in a way so as to improve their scores while maintaining their enforced behavior, we make the agents play a set of 2000 negotiation games which are held out, after every 25000 episodes of training. Figure~\ref{fig:learn_cur} shows this performance. From these curves, we observe the following:  
\begin{enumerate}
\item When both agents are trained to be Prosocial, they gradually converge to the same score. This is not surprising as both agents are being trained to reach a \textit{middle ground}.
\item When both agents are trained to be Selfish, one of the agents ends up scoring more than the other. This, we observe is because it is impossible to reach an agreement (which is the ultimate goal) if both agents are stubborn hence it is necessary for at least one agent to \textit{compromise}.
\item When one agent is trained to be Selfish and the other to be Prosocial, the Selfish one clearly outscores the Prosocial one because the selfish agent and will seldom concede. The prosocial agent on the other hand, concedes more easily.
\end{enumerate}

To test whether the RL agent learns something non-trivial, we compare its performance with two simple baselines:
\begin{enumerate}
\item \textit{RANDOM} At every step, the agent chooses random number of bits to be flipped.
\item \textit{COMMON} Agent 1 (agent who goes first) gives its most selfish offer followed by Agent 2 who does the same. At the third step, Agent 1 offers the intersection of the first two offers (common benefit) to which Agent 2 agrees. If there is no intersection, it is a disagreement.
\end{enumerate}
The results in Table~\ref{tab:coord} are an average over a separate test set of 30000 negotiations. It is clear from the results listed out in table that all three variants of behavioral combinations do better than the baselines in terms of Optimality and Joint Reward. This shows that the agents which are trained against each other learn to \textit{coordinate} their moves such that apart from maintaining their enforced behavior, they try to maximize their scores as well as the optimality. Observe that joint reward is maximum when both agents are prosocial as both agents are not only concerned with maximizing their own reward but also of their opponent's so as to reach optimal deals. This corroborates the results reported in Ref.~\cite{cao2018iclr}.

Figure~\ref{fig:freq_dist_seq} shows the frequency distribution of the sequence of actions (an action is the number of bits flipped by an agent in the opponent's offer) taken by the RL agents during the course of negotiations during test time. In each bar, the orange part represents the number of optimal deals. The distributions show that there is a joint preference among agents for certain sequences more than others which is evident by their \textit{skewed} nature.
\begin{table}
  \caption{Coordination between trained agents. In optimality column, the numbers in bracket are the percentages on the agreed deals.}
  \label{tab:coord}
  \resizebox{\columnwidth}{!}{
  \begin{tabular}{ccccccc}
    \toprule
    \begin{tabular}[c]{@{}c@{}}Agent\\ A\end{tabular} & \begin{tabular}[c]{@{}c@{}}Agent \\ B\end{tabular} & \begin{tabular}[c]{@{}c@{}}Dialog\\ Length\end{tabular} & \begin{tabular}[c]{@{}c@{}}Agreement\\ Rate (\%)\end{tabular} & \begin{tabular}[c]{@{}c@{}}Optimality\\ Rate (\%)\end{tabular} & \multicolumn{2}{c}{\begin{tabular}[c]{@{}c@{}}Average\\ Score\end{tabular}} \\
    \midrule
     & & & &  & A(0.70) & B(0.70)\\
     \midrule
     \multicolumn{2}{c}{RANDOM} & 15.90 & 100 & 24.55 & 0.25 & 0.25 \\
     \multicolumn{2}{c}{COMMON} & 3.77 & 79.54 & 70.39 (88.49) & 0.50 & 0.50 \\
     \midrule
     PP & PP & 16.98 & 96.24 & 82.33 (85.55) & 0.65 & 0.66 \\
     SS & SS & 17.47 & 88.31 & 74.88 (84.79) & 0.54 & 0.69 \\
     SP & PS & 13.87 & 91.90 & 86.74 (94.38) & 0.73 & 0.55 \\
  \bottomrule
\end{tabular}
}
\end{table}
This nature of the distributions also raises the question: Do the agents learn to identify a context from their utilities and act according to that or do they learn to give generic outputs irrespective of their utilities? To answer this question, we made the agents negotiate on the test set using only the 20 most frequent sequences. The number on the top of each bar is the percentage optimality using the given sequence throughout the test set.  We observe that none of the numbers are greater than the overall optimality reported in Table~\ref{tab:coord}. This shows that the agents indeed capture the context from their utilities and behave accordingly.

It is interesting to note that the optimality is the highest when both agents are trained to have different behaviors. This, we observe, is because the Prosocial agents make some offers which have negative valued clauses. If both agents are Prosocial, there might be cases where the agents agree on an offer with negative valued clauses for both agents which leads to non optimal solutions. On the other hand, if one of the agents is selfish, these cases will be avoided because the selfish agent won't agree to them. This is illustrated in the Appendix~\ref{app:eg_neg}.

\section{Interplay between Agents}
\label{sec:interplay}
\begin{table}
  \caption{Interplay between Agents. In optimality column, the numbers in bracket are the percentages on the agreed deals. As we get two SSs and PPs after training, we choose the ones which outscore it's opponent during training and use them in the interplay negotiations.}
  \label{tab:interplay}
  \resizebox{\columnwidth}{!}{
  \begin{tabular}{ccccccc}
    \toprule
    \begin{tabular}[c]{@{}c@{}}Agent\\ A\end{tabular} & \begin{tabular}[c]{@{}c@{}}Agent \\ B\end{tabular} & \begin{tabular}[c]{@{}c@{}}Dialog\\ Length\end{tabular} & \begin{tabular}[c]{@{}c@{}}Agreement\\ Rate (\%)\end{tabular} & \begin{tabular}[c]{@{}c@{}}Optimality\\ Rate (\%)\end{tabular} & \multicolumn{2}{c}{\begin{tabular}[c]{@{}c@{}}Average\\ Score\end{tabular}} \\
    \midrule
     & & & &  & A(0.70) & B(0.70)\\
     \midrule
     SP & SS & 26.50 & 59.00 & 55.81 (94.59) & 0.42 & 0.48 \\
     PP & PS & 9.85 & 97.96 & 62.55 (63.85) & 0.51 & 0.68 \\
     PP & SS & 23.98 & 90.01 & 69.80 (77.54) & 0.44 & 0.75 \\
     SP & PP & 24.64 & 90.43 & 64.28 (71.08) & 0.71 & 0.45 \\
     SS & PS & 11.89 & 93.03 & 69.43 (74.63) & 0.70 & 0.50 \\
  \bottomrule
\end{tabular}
}
\end{table}
In the previous section, we analyzed the results when an agent negotiates with the agent against which it is trained. But to analyze the performance of an agent against a policy it has never seen, we run the test negotiations between agents who have never seen each other during training. These negotiations are what we refer to as \textit{Interplay Negotiations}. Table~\ref{tab:interplay} shows the results of these negotiations. These results are an average over the test set of 30000 negotiations.

It is clear from the results that the optimalities of the interplay between agents are not very high which is because these agents have never seen each other during training and thus have not been able to develop their policies accordingly.

Moreover, note that the agreement rate is highest (97.96\%) for negotiation between prosocial agents (PP vs PS) and lowest (59.00\%) for selfish agents.

We also observe that in every case, the selfish agents outscore the prosocial agents which confirms their corresponding behaviors. It is also interesting to note the scores when two agents trained with the same reward signal but trained against different opponents negotiate with each other. 
\begin{enumerate}
\item When a selfish agent trained against a selfish agent (SS) negotiates with a selfish agent trained against a prosocial agent (SP), the former outscores the latter by a margin of 0.06 points. This establishes the fact that SS is more selfish than SP.
\item When a prosocial agent trained against a prosocial agent (PP) negotiates with a prosocial agent trained against a selfish agent (PS), the latter outscores the former by a margin of 0.17 points. This shows that PP has learnt to be more generous than PS.
\end{enumerate}
From this, and from other interplay negotiations in Table \ref{tab:interplay} we observe varying degrees of selfish/prosocial behavior in agents with some agents being more selfish than others. 

To verify the consistency in agent behavior, we have shown the differences in scores (Player A - Player B) for all interplay negotiations in the form of a matrix in Table~\ref{tab:diff_matrix}. Here, each entry is the difference in scores when corresponding agents negotiate. We observe that the differences are in increasing order along every row and decreasing along columns. As the agents are arranged in decreasing order of their selfishness (from left to right and top to bottom), this kind of distribution confirms consistency in their behavior (i.e. if A beats B with a margin $m$ and B beats C, then A should be able to beat C with a margin greater than $m$).

%\begin{table}
%  \setlength{\extrarowheight}{2pt}
%  \begin{tabular}{*{6}{c|}}
%    \multicolumn{2}{c}{} & \multicolumn{4}{c}{Player $B$}\\\cline{3-6}
%    \multicolumn{1}{c}{} &  & $SS$  & $SP$ & $PS$ & $PP$ \\\cline{2-6}
%    \multirow{4}*{Player $A$}  & $SS$ & $-$ & $0.06$ & $0.20$ & $0.31$ \\\cline{2-6}
%    & $SP$ & $-$ & $-$ & $0.18$ & $0.26$ \\\cline{2-6}
%     & $PS$ & $-$ & $-$ & $-$ & $0.17$ \\\cline{2-6}
%      & $PP$ & $-$ & $-$ & $-$ & $-$ \\\cline{2-6}
%  \end{tabular}
%  \caption{Difference in scores for all agent combinations. The selfishness of an agent decreases from left to right and top to bottom}
%  \label{diff_matrix}
%\end{table}

\begin{table}
  \caption{Difference in scores for all agent combinations. The selfishness of an agent decreases from left to right and top to bottom}
  \label{tab:diff_matrix}
  \begin{tabular}{cccccc}
    \toprule
     & & \multicolumn{4}{c}{Player B} \\
      & & SS & SP & PS & PP \\
      \multirow{4}*{Player A} & SS & - & 0.06 & 0.20 & 0.31 \\
      & SP & - & - & 0.18 & 0.26 \\
      & PS & - & - & - & 0.17 \\
      & PP & - & - & - & - \\
  \bottomrule
\end{tabular}
\end{table}

\section{Meta Agent - Modeling Dynamic behavior}
\label{sec:model_mix}
\begin{table}
  \caption{Performance of Meta Agent against all four kinds of agents.}
  \label{tab:meta}
  \begin{tabular}{cccccc}
    \toprule
    \begin{tabular}[c]{@{}c@{}}B\end{tabular} & \begin{tabular}[c]{@{}c@{}}Dialog\\ Length\end{tabular} & \begin{tabular}[c]{@{}c@{}}Agreement\\ Rate (\%)\end{tabular} & \begin{tabular}[c]{@{}c@{}}Optimality\\ Rate (\%)\end{tabular} & \multicolumn{2}{c}{\begin{tabular}[c]{@{}c@{}}Average\\ Score\end{tabular}} \\
    \midrule
     & & &  & Meta & B\\
     \midrule
     PP & 18.68 & 94.41 & 77.15 (81.71) & 0.64 & 0.61 \\
     SS & 19.17 & 86.25 & 73.33 (85.02) & 0.54 & 0.66 \\
     PS & 13.10 & 92.27 & 76.56 (82.97) & 0.69 & 0.55 \\
     SP & 20.53 & 90.22 & 81.40 (90.22) & 0.55 & 0.71 \\
  \bottomrule
\end{tabular}
\end{table}
\begin{figure*}
\includegraphics[width=14cm]{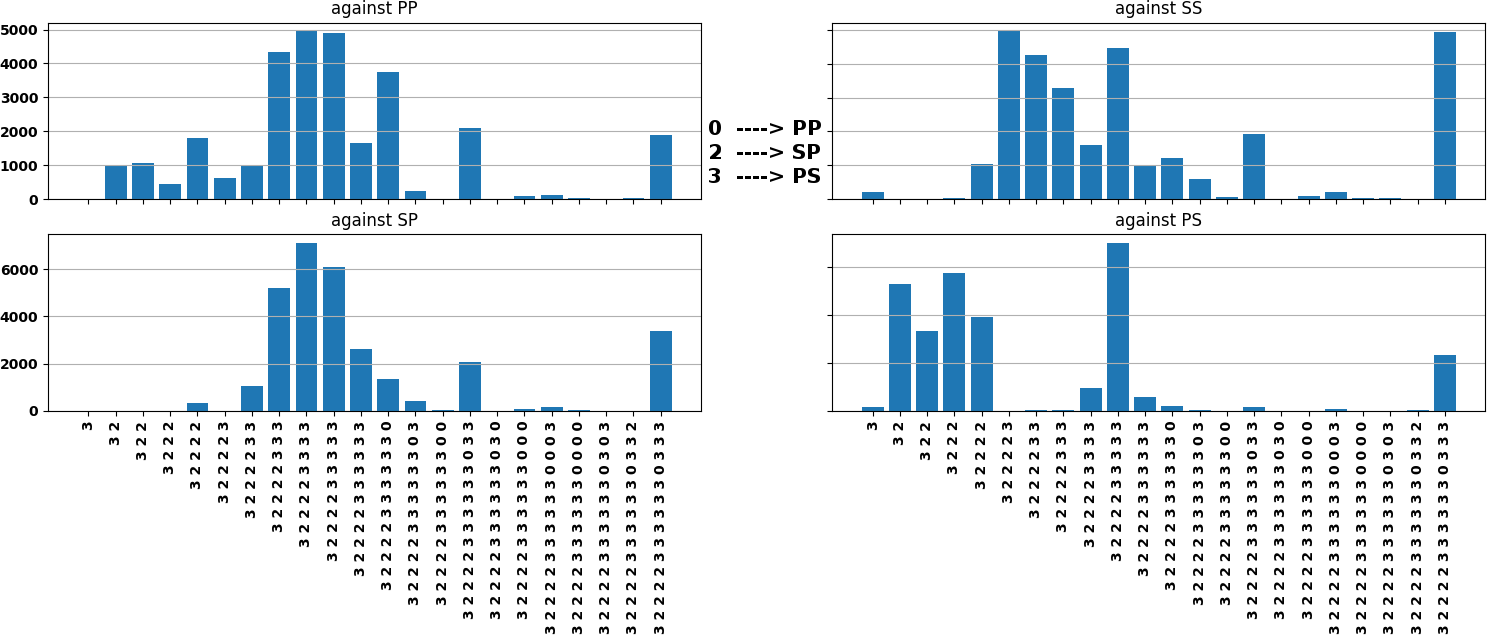}
\caption{Frequency distribution of agent selection by the meta agent. Each point on the x axis represents the agent selection sequence by the meta agent. Each number in this sequence corresponds to a particular agent.}
\label{fig:freq_dist_meta}
\end{figure*}
On the surface, the selfish agent seems to be best as it always outscores its opponents. But using such an agent leads to many disagreements if the opponent is also selfish (see first row in Table~\ref{tab:interplay}). This has also been observed in \cite{guthrie39negotiations} apart from the fact that selfish and prosocial behavior are not, in general, separable processes in negotiation. It is also important to note that humans don't really negotiate using a fixed policy (either prosocial or selfish). They tend to follow a mixed behavior with some degrees of both depending on the state of the negotiation process. According to previous work (\cite{axelrod1981evolution}), there is no universally best policy to negotiate and that it depends on the nature of the opponent. With this motivation, we try to model one policy that works well against all agents by using a mixture of agents. We do this by training another RL agent to choose which of the four agents' policies to use given the state of the negotiation. 
\subsection{Model and Training}
\label{sec:mix_mdl_trn}
The meta agent is an ensemble of all four agents, with a \textit{Selector Agent} on top. The job of the selector agent is to select the output offer of one of the four agents given the context $U$. This agent is modeled as a neural network similar to the one described in Section~\ref{sec:mdl_arch}, except that it also takes the output of all four agents as part of its state input. The selector agent outputs a policy $\pi_s$ from which an action is sampled. This action is the offer produced by one of the four agents. The selector agent maximizes the following objective:
\begin{align}
L_s = \mathop{\mathbb{E}}_{s_t \sim \pi_s}\bigg[\sum_{t}\gamma^{(T-t)}((r_s(s_{1...T})+r_o) - b_s)\bigg] + \lambda H[\pi_s],
\end{align}
where $r_s(s_{1...T})$ is the reward that meta agent gets at the end of the negotiation which is a function of the sequence of actions $s_t$ it takes and $r_o$ is the reward of the opponent. Note that we are giving the joint reward to the meta agent which is a simple way of ensuring that it is not biased towards one particular agent while selecting.

For training, we randomly select one of the four agents as the opponent and make it play a batch of 100 negotiation episodes with the meta agent. During this process, we freeze the weights of the opponent. Similar to Section~\ref{sec:mdl_arch}, we run $10^5$ episodes for 5 epochs.

\subsection{Analysis}
\label{sec:mix_analysis}
We make the meta agent negotiate against each of the four agents one by one on the test set and the results are reported in Table~\ref{tab:meta}. In terms of the scores, the meta agent is able to outscore the prosocial agents but not the selfish ones. The meta agent does well to coordinate with all agents with is reflected by the optimality. Also, the joint reward for all the cases is greater than 1.20. In spite of that, it is not able to match on results reported in Table~\ref{tab:coord}. 

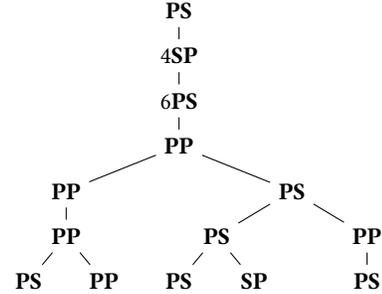
\begin{figure}
\begin{tikzpicture}[level distance=0.6cm,
  level 4/.style={sibling distance=3cm},
  level 5/.style={sibling distance=2cm},
  level 6/.style={sibling distance=1cm}]
  \node {\textbf{PS}}
    child {node {4\textbf{SP}}
      child {node {6\textbf{PS}}
        child{node {\textbf{PP}}
          child{node {\textbf{PP}}
            child{node {\textbf{PP}}
              child{node {\textbf{PS}}
              }
              child{node {\textbf{PP}}
              }
            }
          }
          child{node {\textbf{PS}}
            child{node {\textbf{PS}}
              child{node {\textbf{PS}}
              }
              child{node {\textbf{SP}}
              }
            }
            child{node {\textbf{PP}}
              child{node {\textbf{PS}}}
            }
          }
        }
      }
    };
\end{tikzpicture}
\caption{Decision tree learnt by selector agent. The numbers 4 and 6 at the second and third node denote a repetition of SP and PS 4 and 6 times respectively.} \label{fig:decision_tree}
\end{figure}
\begin{table*}
  \caption{Results of Human Evaluation.}
  \label{tab:human_eval}
  \begin{tabular}{ccccccccc}
    \toprule
    \begin{tabular}[c]{@{}c@{}}Agent\end{tabular} & \begin{tabular}[c]{@{}c@{}}Dialog\\ Length\end{tabular} & \begin{tabular}[c]{@{}c@{}}Agreement\\ Rate (\%)\end{tabular} & \begin{tabular}[c]{@{}c@{}}Optimality\\ Rate (\%)\end{tabular} & \begin{tabular}[c]{@{}c@{}}Agent\\ Score\end{tabular} & \begin{tabular}[c]{@{}c@{}} Human\\ Score\end{tabular} & \begin{tabular}[c]{@{}c@{}}Agent\\ Won (\%)\end{tabular} & \begin{tabular}[c]{@{}c@{}}Human\\ Won (\%)\end{tabular} & \begin{tabular}[c]{@{}c@{}}Tied (\%)\end{tabular} \\
     \midrule
     PP & 15.07 & 87.38 & 70.87 & 0.58 & 0.62 & 36.67 & 51.11 & 12.22 \\
     SS & 19.56 & 73.79 & 60.20 & 0.58 & 0.44 & 60.53 & 21.05 & 18.42 \\
     PS & 13.57 & 92.93 & 66.67 & 0.57 & 0.57 & 40.22 & 52.17 & 7.61 \\
     SP & 21.75 & 72.28 & 59.41 & 0.61 & 0.39 & 68.49 & 20.55 & 10.96 \\
     META & 16.78 & 88.30 & 56.40 & 0.57 & 0.56 & 46.99 & 44.58 & 8.43 \\
     \bottomrule
\end{tabular}
\end{table*}

%\subsubsection*{Selector learns a decision tree}
%\label{dec_tree}
\textit{Selector learns a decision tree:} To analyse what the selector agent actually learns to do, we look at the frequency distribution of the agent selection sequence that it follows against all four agents. This is shown in Figure~\ref{fig:freq_dist_meta}. The interesting part of all the four distributions is the x axis which is the sequence of agent selection. It is same for all four cases. Moreover, every sequence is a subsequence of some larger sequence. This suggests that the selector agent learns to follow a decision tree as shown in Figure~\ref{fig:decision_tree}. Here, every node is the agent selected by the selector at a given state and the edge represents the transition from one state to the next.

The fact that the selector agent learns a decision tree suggests the following:
\begin{enumerate}
\item The agent learns just one policy (the simplest) which works against all agents.
\item We know that it is difficult for an agent to decipher the behavior of the opponent until after a few moves, hence it makes sense to learn just one policy which works well at any stage.
\end{enumerate}

The selector agent starts off with being selfish (SP) and then becomes more generous (PS and PP). This seems like a reasonable policy where an agent tries to maximize its score initially by making selfish offers but starts giving prosocial offers towards the end so as to reach a deal.

Moreover, observe that the selector agent never selects SS as that is the most selfish agent and the chances of reaching a deal with this agent are less (especially against a selfish opponent).

\section{Human Evaluation}
\label{sec:human_eval}

We have established the fact that negotiation agents have learnt to negotiate against each other. But for real life deployment, it is really important to test their performance when negotiating against human players. For this purpose, we organized an experiment in our lab where our colleagues played several rounds of negotiation games with all the five negotiation agents (PP, SS, SP, PS and META). A total of 38 human players negotiated for 3 rounds of negotiation against all 5 agents. This means that each agent played a total of 114 negotiation games against humans. Humans were told that their aim was to maximize their scores. We ensured this by providing them an incentive for every game they outscore the agent (reward). Table~\ref{tab:human_eval} shows the result of the human evaluation.

It is clear from the results that both the selfish agents (SS and SP) outscore humans most of the time. The prosocial agents (PP and PS) on the other hand, get outscored on more occasions. This shows that the behavior of human players is between prosocial and selfish i.e. they have a hybrid behavior which we had hypothesized in Section \ref{sec:model_mix} and that had motivated us build a meta agent. With the meta agent, humans win an almost equal number of times as the meta agent. This proves that we have been somewhat successful in emulating human behavior through our meta agent.

The optimality is not on the higher side. This, we argue, is because humans had no incentive to reach an optimal solution which is the case most of the time in real life. We have argued in Section~\ref{sec:coord} that coordination is necessary to reach optimal solutions, which in turn is attained if the negotiating parties develop their policies according to their opponent.
This opens up an option for future work where AI agents actually undergo learning in real time by playing many negotiating games against humans. The issue, of course, is the scale of the training instances, which needs to be solved.

\section{Conclusion and Future Work}
\label{sec:conc_fw}
We have introduced a simple way of using \textit{Deep Reinforcement Learning} to train agents for contract negotiation. As this is a multiagent setting, we have been successful in modeling varying behaviors in agents by varying the reward signal in the player and its opponent. Empirical evidence shows coordination in agents so as to reach optimal solutions. We have also built a Meta Agent that shows a dynamic behavior by learning an ensemble of models. That our agents can be deployed in real world scenarios is evidenced by the consistency in their trained behaviors against real human players. The fact that we use reinforcement learning to build agents with multiple negotiating behaviors and provide evidence for their usability with human evaluation makes this work a unique contribution. There is much scope for improvement in terms of behavior modeling wherein we may use Reinforcement Learning to learn hyperparameters involved in the \textit{proposal curves} as used in Ref.~\cite{cao2015automated} and also in the reward signals in Ref.~\cite{peysakhovich2017prosocial}. We need to look into ways to train agents to ground their communication in natural language (NL) while negotiating by making them perform a parallel NL task as done in Ref.~\cite{lazaridou2016multi}.

%\end{document}  % This is where a 'short' article might terminate

%\clearpage 
\bibliographystyle{ACM-Reference-Format}
\bibliography{sample-bibliography}
\appendix
\section{Example Negotiation}
\label{app:eg_neg}
An example where agents learn to coordinate and complement their moves is shown in Table~\ref{tab:coord_compl_eg}. This is the actual transcript where we show two different pairs of agents reaching an agreement on the same set of utilities. The top part is when both agents are \textit{PP} and the bottom part is when one agent to be \textit{SP} and the other is \textit{PS}. Here clause 4 is bad for both agents, but when both agents are Prosocial, it finds its way into the final cut. But this doesn't happen when both agents have different behaviors.

\begin{table}
  \caption{Example Negotiation. Top part is the case when both agents are PP. Bottom part shows negotiation on same utilities but when A is SP and B is PS.}
  \label{tab:coord_compl_eg}
  \begin{tabular}{cccccccc}
    \toprule
      & Agent & $C_{1}$ & $C_{2}$ & $C_{3}$ & $C_{4}$ & $C_{5}$ & $C_{6}$ \\
     \midrule
      \multirow{2}*{Utilities} & $A$ & -6 & 12 & -1 & -1 & -3 & -1 \\
      & $B$ & -2 & -6 & -1 & -1 & -2 & 12 \\
     \midrule
      \multirow{8}*{Offers} & $A_{PP}$ & 0 & 1 & 0 & 0 & 0 & 1 \\
      & $B_{PP}$ & 0 & 0 & 0 & 1 & 0 & 1 \\
      & $A_{PP}$ & 0 & 1 & 0 & 1 & 0 & 0 \\
      & $B_{PP}$ & 0 & 0 & 0 & 1 & 0 & 1 \\
      & $A_{PP}$ & 0 & 1 & 0 & 1 & 0 & 0 \\
      & $B_{PP}$ & 0 & 0 & 0 & 1 & 0 & 1 \\
      & $A_{PP}$ & 0 & 1 & 0 & 1 & 0 & 1 \\
      & $B_{PP}$ & 0 & 1 & 0 & 1 & 0 & 1 \\
     \midrule
     Result & \multicolumn{7}{c}{Non optimal solution} \\ %due to inclusion of $C_4$} \\
     \midrule
      \multirow{8}*{Offers} & $A_{SP}$ & 0 & 1 & 0 & 1 & 0 & 1 \\
      & $B_{PS}$ & 1 & 0 & 1 & 0 & 1 & 1 \\
      & $A_{SP}$ & 0 & 1 & 1 & 0 & 0 & 1 \\
      & $B_{PS}$ & 1 & 0 & 0 & 1 & 1 & 1 \\
      & $A_{SP}$ & 0 & 1 & 0 & 1 & 0 & 1 \\
      & $B_{PS}$ & 1 & 0 & 1 & 0 & 1 & 1 \\
      & $A_{SP}$ & 0 & 1 & 1 & 0 & 0 & 1 \\
      & $B_{PS}$ & 0 & 0 & 0 & 1 & 0 & 1 \\
      & $A_{SP}$ & 0 & 1 & 0 & 0 & 0 & 0 \\
      & $B_{PS}$ & 0 & 1 & 0 & 0 & 0 & 1 \\
      & $A_{SP}$ & 0 & 1 & 0 & 0 & 0 & 1 \\
     \midrule
     Result & \multicolumn{7}{c}{Optimal solution} \\ %due to inclusion of $C_4$} \\
     \bottomrule
  \end{tabular}
\end{table}

\section{Hyperparameter settings}
\label{app:hyp}
The $\textrm{OfferMLP}()$ is a 2-layer MLP with $64$ hidden units in each layer and ReLU activation. Both $\textrm{AgentLookup}()$ and $\textrm{TurnLookup}()$ are embeddings with size $32$. The $\textrm{GRU}()$ has a hidden state size of $256$. The baseline $b_i$ and $b_s$ is computed as the running average of the rewards received for each agent $i$. The entropy regularization weight parameter $\lambda$ is changed in every epoch. For $5$ epochs, the values of $\lambda$ are $[0.1, 0.05, 0.01, 0.005, 0.001]$. The discount factor $\gamma$ is set to 0.99.

For training, we used $SGD$ optimizer with nesterov \cite{nesterov1983method} and momentum of $0.1$. The learning rate was fixed to $0.01$. 

\end{document}